%% file: main.tex
\documentclass[conference]{IEEEtran}
\IEEEoverridecommandlockouts
\usepackage{cite}
\usepackage{amsmath,amssymb,amsfonts}
\usepackage{algorithmic}
\usepackage{graphicx}
\usepackage{textcomp}
\usepackage{xcolor}
\usepackage{hyperref}
\usepackage{soul}
\usepackage{caption}
\usepackage{subcaption}
\usepackage[export]{adjustbox}

\usepackage{etoolbox}

\def\BibTeX{{\rm B\kern-.05em{\sc i\kern-.025em b}\kern-.08em
    T\kern-.1667em\lower.7ex\hbox{E}\kern-.125emX}}

\makeatletter
\newcommand{\newlineauthors}{%
  \end{@IEEEauthorhalign}\hfill\mbox{}\par
  \mbox{}\hfill\begin{@IEEEauthorhalign}
}
\makeatother

\title{Explainable Anomaly Detection: Counterfactual driven What-If Analysis}

\author{
    \IEEEauthorblockN{
        Logan Cummins\IEEEauthorrefmark{1}\IEEEauthorrefmark{3},
        Alexander Sommers\IEEEauthorrefmark{1},
        Sudip Mittal\IEEEauthorrefmark{1},
        Shahram Rahimi\IEEEauthorrefmark{1},\\
        Maria Seale\IEEEauthorrefmark{2},
        Joseph Jaboure\IEEEauthorrefmark{2},
        Thomas Arnold\IEEEauthorrefmark{2}
    }
    \IEEEauthorblockA{
        \IEEEauthorrefmark{1}\textit{Computer Science \& Engineering,}
        \textit{Mississippi State University}
        \\\{nlc123, ams1988\}@msstate.edu, \{mittal, rahimi\}@cse.msstate.edu
    }
    \IEEEauthorblockA{
        \IEEEauthorrefmark{2}\textit{Engineer Research and Development Center, } 
        \emph{US Department of Defense}
        \\\{maria.a.seale, joseph.e.jabour, thomas.l.arnold\}@erdc.dren.mil
    }
    \IEEEauthorrefmark{3}corresponding author
}

\begin{document}
\maketitle
\begin{abstract}
There exists three main areas of study inside of the field of predictive maintenance: anomaly detection, fault diagnosis, and remaining useful life prediction. 
Notably, anomaly detection alerts the stakeholder that an anomaly is occurring. 
This raises two fundamental questions: what is causing the fault and how can we fix it?
Inside of the field of explainable artificial intelligence, counterfactual explanations can give that information in the form of what changes to make to put the data point into the opposing class, in this case ``healthy."
The suggestions are not always actionable which may raise the interest in asking ``what if we do this instead?"
In this work, we provide a proof of concept for utilizing counterfactual explanations as what-if analysis. 
We perform this on the PRONOSTIA dataset with a temporal convolutional network as the anomaly detector. 
Our method presents the counterfactuals in the form of a what-if analysis for this base problem to inspire future work for more complex systems and scenarios.
\end{abstract}
\begin{IEEEkeywords}
Temporal Convolutional Networks, Explainable Artificial Intelligence, Counterfactuals, XAI, Predictive Maintenance, Machinery, Explainable Predictive Maintenance, What-If Analysis
\end{IEEEkeywords}

\section{Introduction}\label{sec:introduction}
\input{SECTIONS/introduction}

\section{Background and Related Work}\label{sec:background}
In this section, we describe many topics that must be understood before continuing with our work including what-if analysis, counterfactual explanations and temporal convolutional networks (TCN). 
\input{SECTIONS/whatif_background}
\input{SECTIONS/counterfactual_background}
\input{SECTIONS/TCN_background}

\section{What-If Analysis for Anomaly Detection}\label{sec:method}
In this section, we describe the methodology that we utilize for what-if analysis for anomaly detection. Firstly, we discuss the PRONOSTIA fault bearing dataset and how we handle class imbalance. Next, we describe the two experiments done which includes k-fold cross validation to validate the TCN architecture and full model training to see its performance on the PRONOSTIA dataset. We then describe the process of counterfactual explanation generation through CoMTE. Lastly, we describe how we use these counterfactuals as an online what-if analysis method for anomaly detection. 
\input{SECTIONS/architecure_diagram_v2}
\input{SECTIONS/pronostia}
\subsection{TCN for Anomaly Detection} \label{sec:ML4AD}
For anomaly detection, we apply the Temporal Convolution Network (TCN). 
As stated in Section \ref{sec:background}, the TCN architecture has seen much success for time series related tasks including PdM related tasks \cite{jakubowski2022roll,Li2021-sb, Cao2021-ij, chen2022utrad, neupane2023twinexplainer}.
For both of our experiments described below, we utilize the hyperparameters listed in Table \ref{tab:hyperparameters}. 

\input{SECTIONS/TCN_hyperparameters}
\subsubsection{K-Fold Cross Validation} \label{sec:method_kfold}
\input{SECTIONS/kfold_exp}
\subsubsection{Full Model Training} \label{sec:method_model_training}
\input{SECTIONS/TCN_exp}
\subsection{Offline Counterfactual Generation} \label{sec:method_comte}
\input{SECTIONS/Comte}
\subsection{Online What-if Analysis} \label{sec:method_whatif}
\input{SECTIONS/whatif_method}

\section{Results}\label{sec:results}
In this section, we present our models' testing results. 
These results of our k-fold cross-validation experiment can be seen in Table \ref{tab:kfold}. 
The results of our full anomaly detection experiment can be seen in Table \ref{tab:tcn_performance}.
\subsection{K-Fold Cross-Validation Results}
\input{SECTIONS/results_kfold}
\subsection{TCN Performance Results}
\input{SECTIONS/results_tcn}

\section{An Illustration of What-if Analysis for Anomaly Detection}\label{sec:illustration}
In this section, we describe a use-case for utilizing counterfactuals explanations as what-if analysis.
This use-case relies on and follows the scenario described by Fig. \ref{fig:storyboard}.

\section{Discussion} \label{sec:dis}
In this section, we discuss the larger message from our experiment as well as the use-case presented. 
This describes a number of important aspects about our model and method chosen as well as how this could be applied in different settings. 
\input{SECTIONS/discussion}

\section{Limitations and Future Work} \label{sec:limit}
In this section, we describe the limitations of both the individual elements we used for testing such as the model and the dataset. 
Additionally we describe the limitations of this specific counterfactual explanation generation approach. 
\input{SECTIONS/limitation}

\section{Conclusions}\label{sec:conclusion}
\input{SECTIONS/conclusion}

\section{Acknowledgement}
This work by the Mississippi State University was financially supported by the U.S. Department of Defense (DoD) High Performance Computing Modernization Program, through the US Army Engineering Research and Development Center (ERDC) (\#W912HZ21C0014). The views and conclusions contained herein are those of the authors and should not be interpreted as necessarily representing the official policies or endorsements, either expressed or implied, of the U.S. Army ERDC or the U.S. DoD. Authors would also like to thank Mississippi State University's Predictive Analytics and Technology Integration (PATENT) Laboratory for its support.

\bibliographystyle{IEEEtran}
\bibliography{references}
\end{document}

%% file: SECTIONS/introduction.tex
\newbool{AV}
\setbool{AV}{true}
\ifbool{AV}
{
    Downtime is feared by utility companies, internet service providers, system administrators, and production lines of all kinds. A system or service being down threatens loss of consumer confidence and contracts. 
    For this reason and more, stakeholders are eager to minimize downtime while acknowledging the need for maintenance related downtime.
    
    Condition based maintenance (CBM) seeks to thread the needle of expense between preventative (before needed) maintenance and the often greater-expense of reactive (after needed) maintenance, intervening exactly as needed. Employing machine learning (ML) to accomplish CBM constitutes predictive maintenance (PdM) \cite{cummins2024explainable}. PdM employing powerful data-driven models  can learn highly non-linear and complex functions between sensible conditions and the incipience of some breakdown. However, many of these models are black-boxes which requires recommendations to be taken on faith. This is often unacceptable, and explainable AI (XAI) methods can satisfy the need to check the model's work \cite{cummins2024explainable}.

    Anomaly detection (AD) is a key task supporting PdM \cite{Carletti2023-ck, Beretta2021-fq, steenwinckel2021flags, marcato2021machine, Gribbestad2021-wq}. Here, AD classes signals indicative of malfunction as anomalous, and others as nominal \cite{cummins2024explainable}. In the context of AD, the XAI method of counterfactual explanation generation can try to answer, for a detected anomaly, ``What is anomalous here? Why? \emph{What if} conditions were different?" \cite{cummins2024explainable}. This facilitates the practice of \emph{what-if} analysis, with two broad uses. 
    
    The first use interrogates an anomaly's cause. Given an anomalous point in the space of sensor inputs, it may be that reduction in this vibration frequency, those temperatures, or that voltage, would return the system to a nominal state. By perturbing the anomalous point, discovering the change which most readily returns it to the nominal class, an AD system can suggest loci of malfunction. If reducing a particular voltage easily reclassifies the point as nominal, then subsystems responsible for that voltage are implicated as malfunctioning. 
    
    The second use suggests supplementary action. The obvious use of localized malfunction is to investigate, and possibly repair, implicated subsystems. But this is not always possible. Suppose a broken cooling system in a generator has allowed a circuit board to become dangerously hot. What-if analysis has already implicated the heat pump, but no replacement is available, but what-if analysis remains useful. Since it has localized the malfunction not only to a set of subsystems (the heat pump) but to a condition (the reduction of temperature), the supplementary action of reducing load on the generator may allow the damaged heat pump to compensate adequately until a replacement can be found. In practice, a typical sequence using this technology is illustrated in Fig. \ref{fig:storyboard}. In Fig. \ref{fig:sub1}, a detected anomaly is interrogated using what-if analysis. The user interprets the results, realizes that the obvious fix cannot be implemented immediately, and uses what-if analysis again to search for a condition based countermeasure (Fig. \ref{fig:sub2}). These results are interpreted, the user sees their planned systemic modification may counteract the anomalous condition (Fig. \ref{fig:sub3}). 
    
    \input{SECTIONS/storyboard}

    The present work contributes a proof of concept implementation of counterfactual generation for AD what-if analysis on multivariate data, with the work organized as follows. Section \ref{sec:background} supplies concepts prerequisite to understanding the work, and summarizes the prior-work motivating the present. Section \ref{sec:method} presents the developed what-if analysis framework, and the experiments used to test it. Sections \ref{sec:results} and \ref{sec:dis} present the results, and a discussion of their implications, respectively. Section \ref{sec:illustration} illustrates a use case of what-if analysis for anomaly detection. Finally, section \ref{sec:limit} describes the limitations in our work, and section \ref{sec:conclusion} concludes with a summary of the work, and an outline of future directions.
    
}
{
    In the realm of mechanical systems, downtime mitigation provides an obvious benefit to the stakeholders. 
    There have been many approaches utilized for mitigating downtime; however, several recent approaches exist in the domain of predictive maintenance (PdM) \cite{cummins2024explainable}. 
    PdM allows experts to use popular artificial intelligence (AI) methods to predict when a maintenance event will need to occur. This predictive power can allow for planned maintenance events as opposed to a reactionary approach that would have an actual component break and bring down a system.
    A popular problem within PdM is anomaly detection. 
    
    Anomaly detection simply aims to learn the difference between healthy and anomalous data \cite{cummins2024explainable}.
    There are numerous approaches to perform this task from tree-structures \cite{Carletti2023-ck, Beretta2021-fq} to deep learning \cite{steenwinckel2021flags, marcato2021machine, Gribbestad2021-wq}.
    However, these methods are not necessarily understandable to the stakeholders that would be performing the maintenance on these systems \cite{cummins2024explainable}. This highlights the importance of introduction of explainable AI (XAI) into the field of PdM. 
    
    XAI bridges the gap between users and AI/ML systems by employing meaningful and explainable methods. As Cummins et al. shows in \cite{cummins2024explainable}, different methods of XAI have been applied to predictive maintenance, such as Shapley Additive Explanations, Local Interpretable Model-agnostic Explanations and more. Among these methods, counterfactual explanation generation stands out as a unique approach that uses perturbations to alter the actual class label of the explained data point.
    This perturbation reveals the important features of the prediction, their behavior, and how these features might have appeared if their behavior had been different.
    From this mental framework, one could see counterfactuals as a method of what-if analysis where the ``what-if" question is framed as ``what would the data look like if the system had resulted in a different class label?"
    From a different perspective, the question could be framed as ``what if we make a change to the system (then will our system be healthy)?"
    Consequently, this framework would allow us to answer at least two types of what-if questions for explainable PdM systems. 
    
    This idea can be illustrated for anomaly detection in Fig. \ref{fig:storyboard}.
    A black-box PdM system would alert a user of a detected anomaly. 
    Through an interactive system, the user would question the system for an explanation of the anomaly as seen in  Fig. \ref{fig:sub1}.
    The user may not always be able to fix that exact component at that given time. 
    A counterfactual could be generate to see what could be changed to put the system in a healthy state.
    From the user's perspective, they would be asking a ``what-if" question typically in the form of ``what if I did this instead?"
    This questioning can be seen in Fig. \ref{fig:sub2}.
    If the system were able to come back with a counterfactual, this would perform the equivalent of a what-if analysis from the user's perspective where they could take an alternate course of action, as seen in Fig. \ref{fig:sub3}.
    They would now be able to move their system out of an anomalous state without tackling the main component causing an anomaly.
    

\input{SECTIONS/storyboard}
    This article's main contribution is utilizing counterfactuals to explain multivariate anomaly detection and provide the explanations in the framework of what-if analysis.
    The rest of the article is organized into the following sections. 
    Section \ref{sec:background} describes much of the information that is needed to understand our approach.
    Additionally, it summarizes some recent works that motivate ours. 
    Section \ref{sec:method} describes framework that we are utilizing for what-if analysis applied to anomaly detection.
    Additionally, this section describes the details associated with our experiments. 
    Section \ref{sec:results} describes the results of our experimentation while Section \ref{sec:dis} discusses the findings and implications.
    Section \ref{sec:illustration} illustrates a use case of what-if analysis for anomaly detection. 
    Finally, section \ref{sec:limit} describes the limitations in our work, and section \ref{sec:conclusion} concludes our article by summarizing our work and outlining future directions.
}

%% file: SECTIONS/storyboard.tex
\begin{figure*}
\centering
\begin{subfigure}{.32\textwidth}
  \centering
  \includegraphics[width=0.9\linewidth]{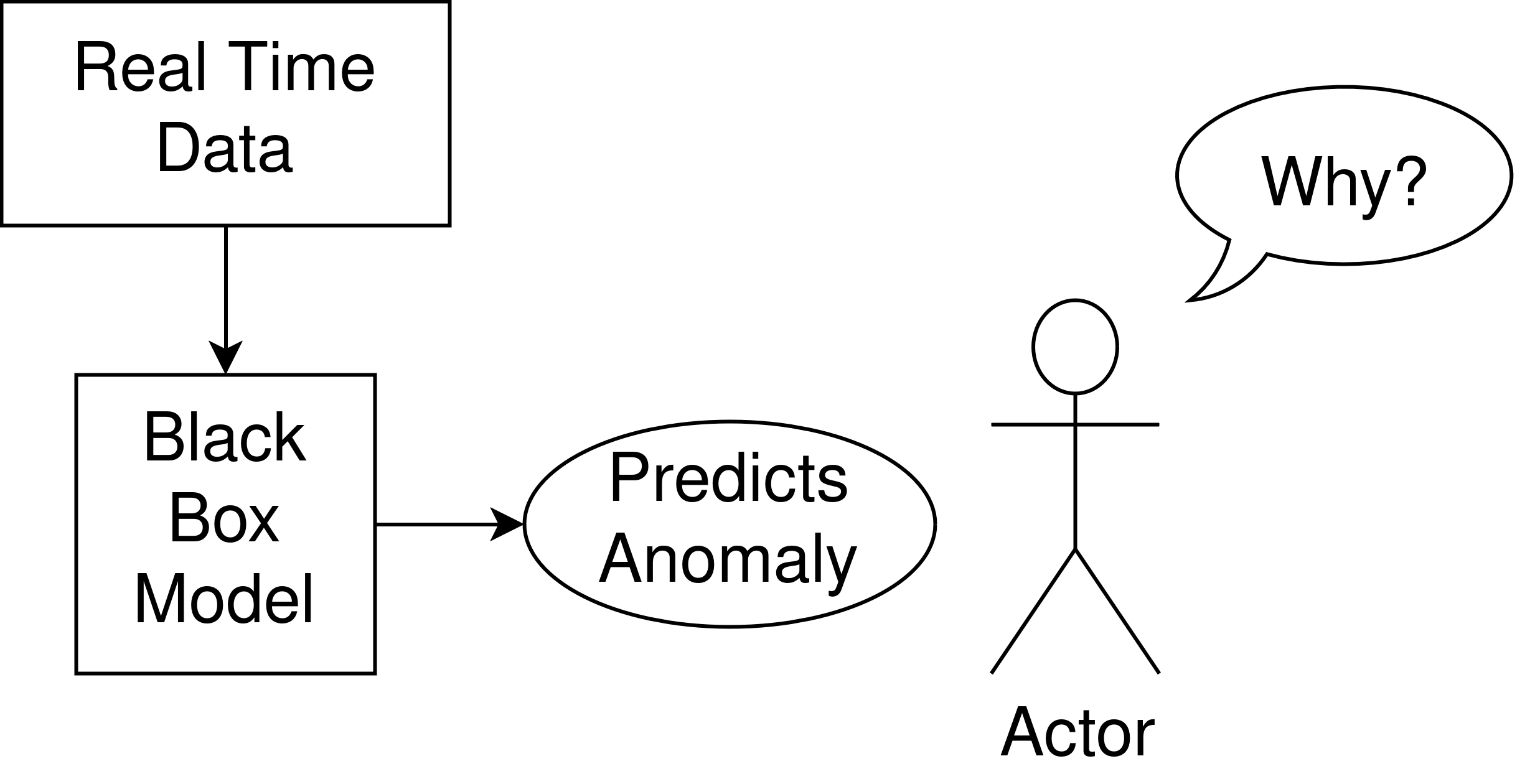}
  \caption{Anomaly detected and user prompts for an explanation}
  \label{fig:sub1}
\end{subfigure}%
\unskip\ \vrule\
\begin{subfigure}{.32\textwidth}
  \centering
  \includegraphics[width=\linewidth]{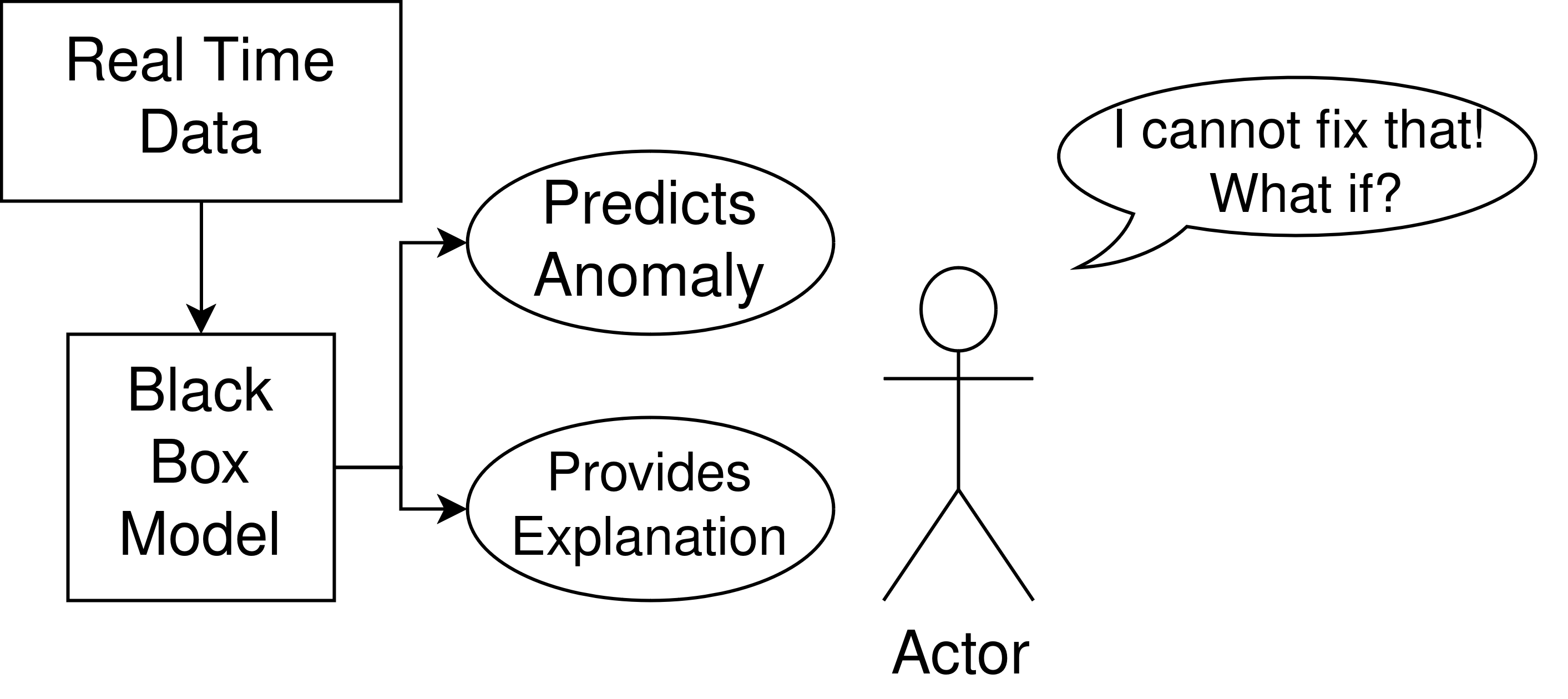}
  \caption{User is unsatisfied and suggests a different fix}
  \label{fig:sub2}
\end{subfigure}%
\unskip\ \vrule\
\begin{subfigure}{.32\textwidth}
  \centering
  \includegraphics[width=\linewidth]{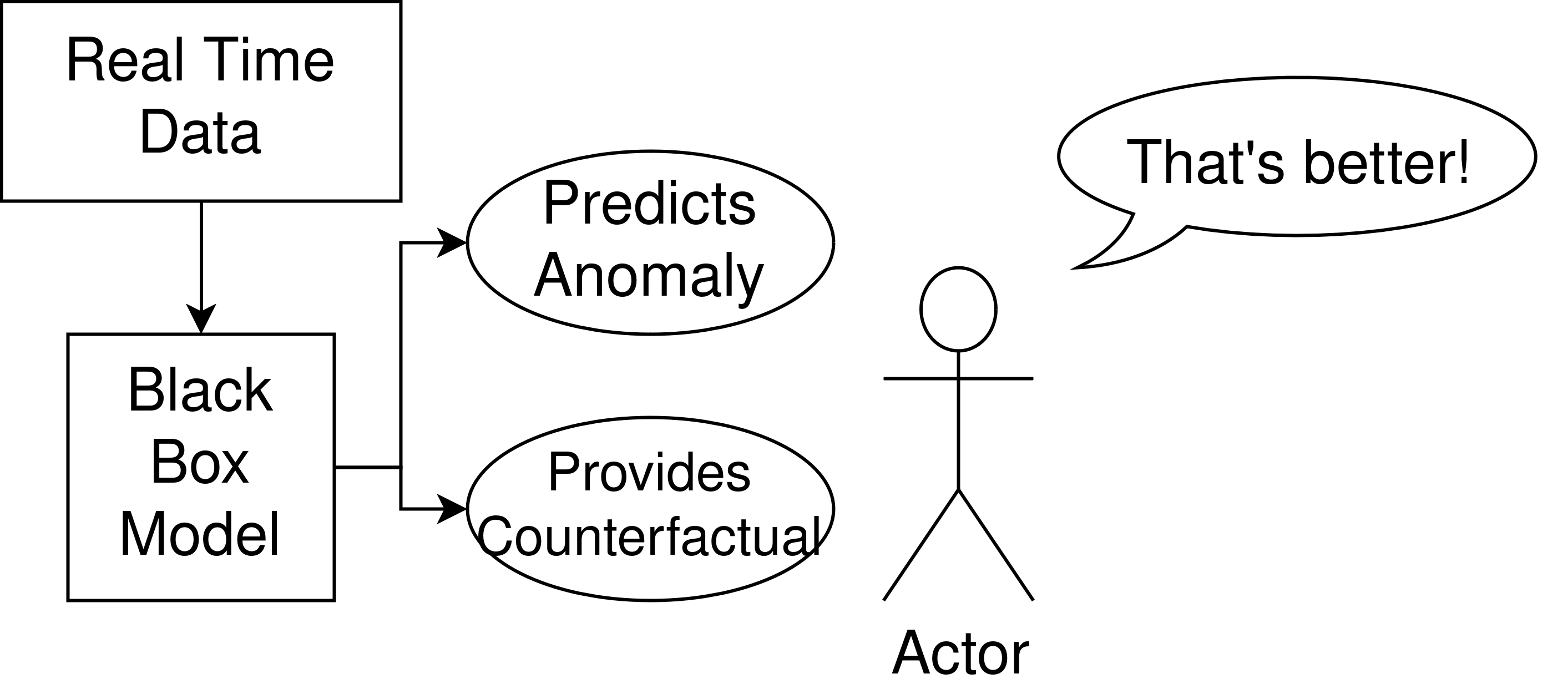}
  \caption{User receives what would happen given the suggested fix}
  \label{fig:sub3}
\end{subfigure}
\caption{Story board of counterfactuals as a what-if analysis. The user received a notification of an anomaly (a). The user may then receive an unactionable explanation (b). The user could then perform what-if analysis via counterfactuals to test a more actionable explanation (c).}
\label{fig:storyboard}
\end{figure*}

%% file: SECTIONS/whatif_background.tex
\subsection{What-if Analysis}
\begin{figure}[h]
    \centering
    \includegraphics[width=0.51\textwidth]{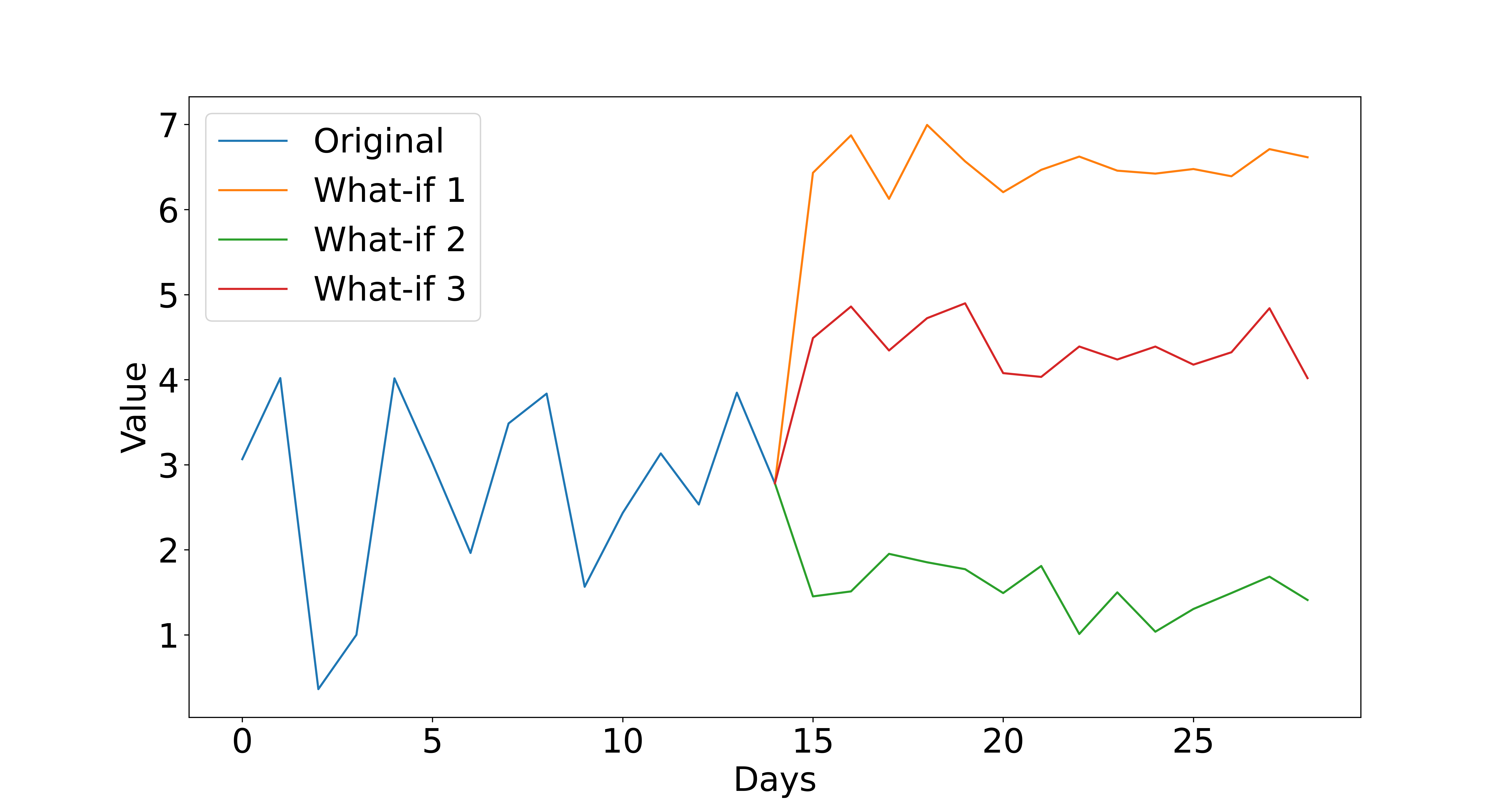}
    \caption{Example of what-if analysis for the value of an item over time}
    \label{fig:what-if}
\end{figure}

Golfarelli et al. \cite{Golfarelli2006-dj} describe what-if analysis as a ``data-intensive simulation whose goal is to inspect the behavior of a complex system under some given hypothesis."
More succinctly, a what-if analysis answers the question ``how do changes in X impact Y?" 
Figure \ref{fig:what-if}, shows an example of what-if analysis used in prediction over a number of days. 
The known value can be seen with three diverging paths.
These three paths would resemble different what-if analyses with different changes impacting the value. 
Additionally, what-if analysis can be performed in many different ways; a few are provided below.

Dandolo et al. \cite{dandolo2023acme} propose Accelerated Model-agnostic Explanations (AcME) as a global and local explainable AI method.
This method perturbs each feature of the dataset on the basis of the quantiles found in the distribution of the feature. 
This allows for less calculations than other methods such as SHAP and LIME. Additionally, this explainable method can be used for what-if analysis by perturbing the data point to fit the query and showing the output of the model.
Singh et al. \cite{singh2013analytical} propose a what-if analysis system for cloud computing applications.
Their main what-if analysis pipeline consists of three parts: finding the components that are impacted by the what-if query, modeling the components in a directed acyclic graph (DAG), and propagating the change proposed in the what-if query through this DAG to see the influence.
Their method shows promising results in monitoring and accurately predicting the different resources, such as CPU utilization and workload, that would be used for cloud computing applications.

Herodotou and Babu \cite{herodotou2011profiling} propose a numerous additions to a big data analytics system, including what-if analysis that works in two parts.
Firstly a profile is generated that represents a hypothetical job of their systems. 
This profile is represented as a dataflow statistics and the costs associated with the dataflow. 
These statistics are proportional to the input data that would be expected from a typical job or from a job that is similar from their seen data. 
With the different statistics, they are able to simulate the job using white-box models; thus simulating a situation that has not happened. 

%% file: SECTIONS/counterfactual_background.tex
\subsection{Counterfactual Explanation Generation}
Counterfactual explanation generation for XAI was introduced by Wachter et al. \cite{Wachter2017-pi} as a way of providing a difference between a data point and a desired outcome \cite{cummins2024explainable}.
Counterfactual explanation generation aims to minimize the amount of changes necessary to a data point while achieving a change in model output. 
This is done a number of different ways, but it can be generalized into the form
\begin{equation}
   \arg\min_{V'}\max_\lambda \lambda(f_w(V') - p')^2 + d(V_i, V') 
\end{equation}
where $V'$ is a counterfactual that is similar to the original data vector $V_i$, $f_w(V')$ is the output of the model being fed the counterfactual, $p'$ is the desired target, $d(*,*)$ is a distance function that measures how difference between the counterfactual and the original datapoint, and $\lambda$ is the objective function \cite{Wachter2017-pi}. This could also be put into a verbal representation similar to \emph{classification $p$ was returned. If variables $V$ had different values ($v_1, v_2,...$), and all other variables had remained constant, classification $p'$ would have been returned} \cite{Wachter2017-pi}. 

Counterfactuals have been utilized as a way of explaining the output of time series modeling tasks. 
Jakubowski et al. \cite{jakubowski2022roll} incorporated counterfactuals to explain their Physics-based Autoencoder architecture. 
First, they only search the test dataset for a suboptimal yet valid counterfactuals. 
Then they search the domain around that data point for the best fit counterfactual. 
The main limitation from their approach, as stated in their work, was the counterfactual generation speed that would not suffice for real-time explanation generation.

Ates et al. \cite{ates2021counterfactual} proposed Counterfactuals for Multivariate Time-series Explanations (CoMTE).
Their method aims to find a fast solution, to combat the NP-Hard problem of optimal counterfactual generation.
The speed comes from their usage of a \textit{k-d tree}, a space partitioning tree introduced by \cite{Bentley1975-dt}. 
A k-d tree helps them compute the most similar data point in the training dataset (the distractor) to the test point.
Their method iteratively replace features from the distractor to the test point until the prediction has changed.
We utilize this method in this work, and describe it in more detail in Section \ref{sec:method}.

Lastly, Lang et al. \cite{lang2023generating} proposed a Generative Adversarial Network (GAN) based approach for counterfactual generation known as SPARse Counterfactual Explanations (SPARCE).
To ensure the realism of the counterfactual, they create a GAN to learn the distributions of the data and produce counterfactuals that mimic the distributions.
Since they are using temporal data, their GAN is implemented as a set of bidirectional LSTMs. 
They tested their method on three different datasets, notably no anomaly detection datasets, and verified that their approach performs the best or second best in terms of precision, similarity, sparsity, and plausibility.  

%% file: SECTIONS/TCN_background.tex
\subsection{Temporal Convolutional Network (TCN)}
Proposed in 2018 by Bai et al. \cite{bai2018empirical}, the TCN is a generic architecture for convolutional sequence modeling that was designed to compete with the best recurrent models. 
This architecture is based on two principles: input-output length homogeneity and no information leakage \cite{bai2018empirical}. 
The first principle is achieved by keeping layer lengths consistent throughout the network. 
The second principle is achieved via their causal convolution mechanism which determines the output at time $t$ by using only the data from $t$ and earlier. 
Additionally to handle long history sizes, they adopt two mechanisms from convolutional architectures, dialated convolution and residual connections. 
Dialated convolution allows them to increase the size of the receptive field while decreasing the amount of computation by logarithmically decreasing the amount of elements included between computation layers.
Residual connections are used to connect the output of the dialated convolutions and the original input which promotes learning of mappings rather than transformations \cite{bai2018empirical}.

TCNs have been applied to predictive maintenance in the forms of anomaly detection and remaining useful life prediction.
He et al. \cite{He2019-ek} performed anomaly detection on three different datasets, electrocardiograms, 2-D gesture and space shuttle.
Additionally, they created two TCN architectures, one base TCN architecture and one with the addition of branching from different layers to measure multiple scales of the same features. 
They found that TCN were were quite successful in performing sequence modeling; moreover, their modification performed better than the base model.

TCNs have seen more use in the realm of remaining useful life prediction.
Li et al. \cite{Li2021-sb} applied the TCN architecture to the famous CMAPSS dataset \cite{saxena2008turbofan}.
Compared to six other architectures, including other sequence modeling architectures such as LSTMs and DCNNs, the TCN outperformed the competition in both RMSE and Score.
Chen et al. \cite{Chen2021-pj} also applied a base TCN to the CMAPSS dataset. 
They expanded the list of architectures that were compared and showed that generally the TCN performed best or second best to CapsNet.
Wang et al. \cite{Wang2023-bi} improved upon the TCN with the addition of dual competitive attention (DCA) to create the Competitive TCN (CTCN).
The DCA is comprised of three components: global competition which removes features with little impact, dual attention which calculates the attention weights in the temporal and channel dimensions, and multidimensional competition which fuses the two into one dimension.
When applying their approach to RUL prediction of the PRONOSTIA dataset, they found their architecture to perform the best in eleven out of thirteen datasets in comparison to CNN, CNN with DCA, and TCN. 
With the successful use of TCN as a time series modeling approach, we feel comfortable using TCN-based architectures for our task.
Now we describe the framework we utilize to apply what-if anaylsis to anomaly detection with the aid of counterfactuals.

%% file: SECTIONS/architecure_diagram_v2.tex
\begin{figure*}[ht]
    \centering
    \includegraphics[width=0.9\linewidth]{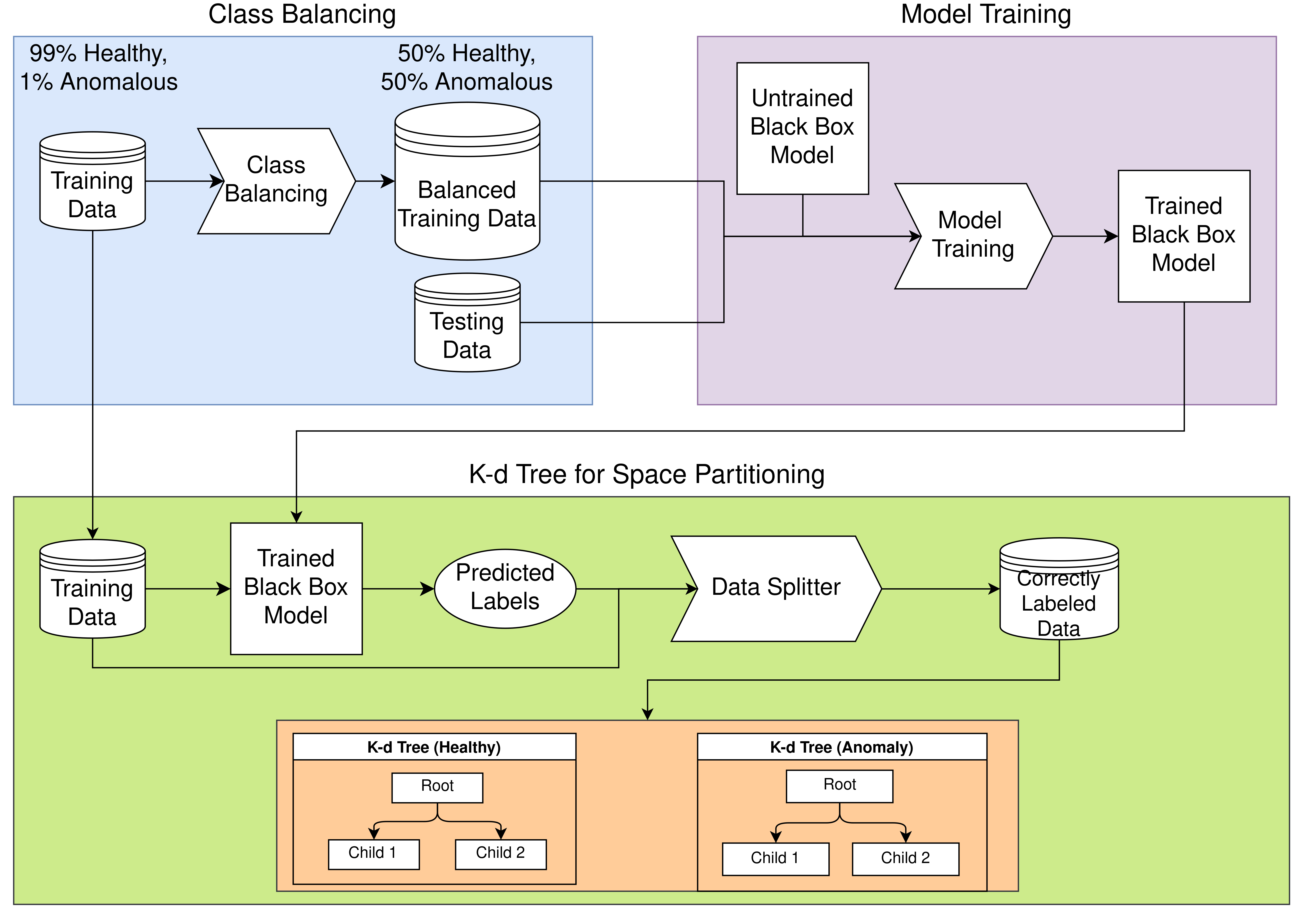}
    \caption{Data flow model beginning with class balancing and ending with k-d tree creation. Class balancing is used to increase the size of the training data to an appropriate ratio of healthy and anomalous data. The model is trained using the balanced data and tested with the original testing data. K-d trees are the created to represent the accurately predicted data.}
    \label{fig:architecture}
\end{figure*}

%% file: SECTIONS/pronostia.tex
\subsection{PRONOSTIA Dataset}
The PRONOSTIA dataset \cite{nectoux2012pronostia} was originally introduced as a real data benchmark dataset for the IEEE PHM 2012 Prognostic Challenge. 
This dataset consists of a learning dataset and a testing dataset split across 3 operating conditions. 
Additionally, the training datasets consist of 2 run-to-failure instances per operating condition. 
The testing datasets consist of 5 instances for operating condition 1, 5 instances for operating condition 2, and 1 instance for operating condition 3. This information can also be seen in Table \ref{tab:dataset}.

\begin{table}[ht]
    \centering
    \begin{tabular}{|c|c|c|c|}
    \hline
    \textbf{Dataset} & \textbf{OC1} & \textbf{OC2} & \textbf{OC3} \\ \hline
    Training & Bearing1\_1 & Bearing2\_1 & Bearing3\_1 \\
            & Bearing1\_2 & Bearing2\_2 & Bearing3\_2 \\ \hline
    Testing & Bearing1\_3 & Bearing2\_3 & Bearing3\_3 \\ 
            & Bearing1\_4 & Bearing2\_4 & \\
            & Bearing1\_5 & Bearing2\_5 & \\
            & Bearing1\_6 & Bearing2\_6 & \\
            & Bearing1\_7 & Bearing2\_7 & \\ \hline
    \end{tabular}
    \caption{PRONOSTIA dataset breakdown}
    \label{tab:dataset}
\end{table}

\begin{figure}
    \centering
    \includegraphics[width=0.45\textwidth]{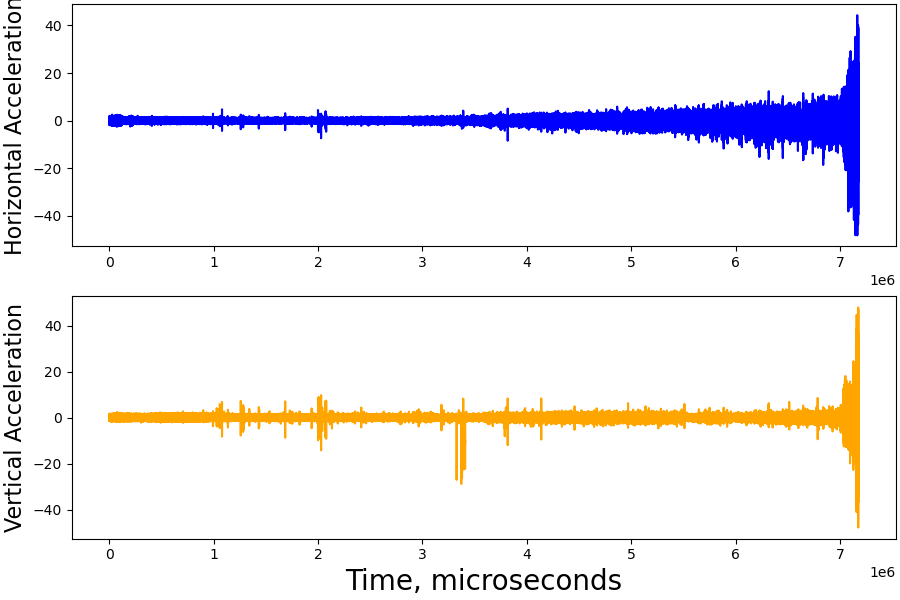}
    \caption{Bearing1\_1 dataset split into its two features}
    \label{fig:bearing1}
\end{figure}

Each dataset contain two variables, a horizontal acceleration and a vertical acceleration as seen in Figure. \ref{fig:bearing1}.
Additionally, there is a temperature value that was not utilized which is not uncommon due to being a different frequency from the acceleration variables. 
For labeling the data set, we utilized the three sigma rule \cite{wang2016two, hu2020anomaly}.
We calculated the root mean square (RMS) of the different features over the 2560 length time vector. 
After calculating the mean and standard deviation of the RMS, we labelled all of the data vectors that are three standard deviations over the mean as anomalous. 
Lastly, to have a smoother labeling scheme, every data point was labeled anomalous if it was followed by only anomalous data points. 

\subsubsection{Class Balancing through SMOTE}
With our class separation criteria, we have a glaring problem: class imbalance. 
This could cause many issues in training such preference to the majority class \cite{Fernandez2018-ld}.
To get past this issue, we perform Synthetic Minority Oversampling Technique (SMOTE) \cite{Chawla2002-ub}. 
This method involves utilizing clustering, in this case k-nearest neighbors clustering, of the minority class, in this case anomalous. 
For each data point in the anomalous class, the difference between that data point and its nearest neighbor is calculated, and the difference is used to slightly perturb the original data point. 
The newly created data point is then added to the dataset. 
Next we describe the architecture used in our experiments. 

%% file: SECTIONS/TCN_hyperparameters.tex
\begin{table}[ht]
    \centering
    \begin{tabular}{|c|c|}
    \hline
    \textbf{Hyperparameter} & \textbf{Value} \\ \hline
        epochs              & 10 \\
        batch size          & 32 \\
        dropout rate        & 0.05 \\
        kernel size         & 7 \\
        levels              & 1 \\
        learning rate       & $2 * 10^{-3}$ \\
        hidden nodes per level & 20 \\ \hline
    \end{tabular}
    \caption{Hyperparameters for TCN models}
    \label{tab:hyperparameters}
\end{table}

%% file: SECTIONS/kfold_exp.tex
Our first experiment aimed to determine if the TCN is truly going to be able to generalize on our dataset \cite{anguita2012k}.
We trained three different models, one for each operating condition.
We utilized a 5-fold cross-validation set-up where each subset of data was loaded, shuffled, and split into five smaller datasets. 
As this architecture requires the data to be split into time-vectors prior to training, we believe this is representative to commonly seen k-fold implementations on non-time series data. 
As per specifying 5-fold, the models were then trained on four of the smaller datasets and tested on the remaining one. 

%% file: SECTIONS/TCN_exp.tex
After performing k-fold cross validation, we trained our mulitple TCN models for our dataset. 
As we have three operating conditions, we trained four models.
The four models are known throughout the paper as $TCN_1, TCN_2, TCN_3,$ and $TCN_{123}$ where the subscript designates what operating conditions were used to train the model.
This allowed us to compare the generalizability of the models trained on their full corresponding datasets. 
Additionally, this allowed us to compare these to a model trained on all three operating conditions. 

For thorough testing of our models, we utilized five metrics that are commonly used in classification problems.
These metrics were accuracy, recall, precision, f-score and g-score \cite{niaz2022class}.
These are all built upon the confusion matrix that shows true positive, true negative, false positive and false negative information. 
Accuracy is the percentage of correct predictions. 
Precision is the percentage of correctly labeled positive results in all positive labeled results. 
Recall is the percentage of correctly labeled positive results. 
F-score shows how accurate the model is on the dataset. 
Lastly, G-mean shows the performance of the model on the majority class and the minority class. 


%% file: SECTIONS/Comte.tex
For the counterfactual generation module of our framework, we utilized the Counterfactuals for Multivariate Time-series Explanation (CoMTE) method proposed by Ates et al. \cite{ates2021counterfactual}.
CoMTE is one of the few methods that allows for counterfactual generation for multivariate time series data.
This method is broken up into two steps that are described below. 

\subsubsection{K-d Tree for Space Partitioning}
CoMTE starts out by choosing \emph{distractor candidates} that serve as the basis of the counterfactuals.
Features that are desirable for the method of gathering the distractor candidates are efficiency and realness. 
Efficiency refers to the time complexity of the method to generate the distractor candidates, and realness refers to the likelihood for the data to exist. 
They provide these two features by utilizing a k-d tree, proposed by Bentley \cite{Bentley1975-dt}

\begin{figure}
    \centering
    \includegraphics[width=0.45\textwidth]{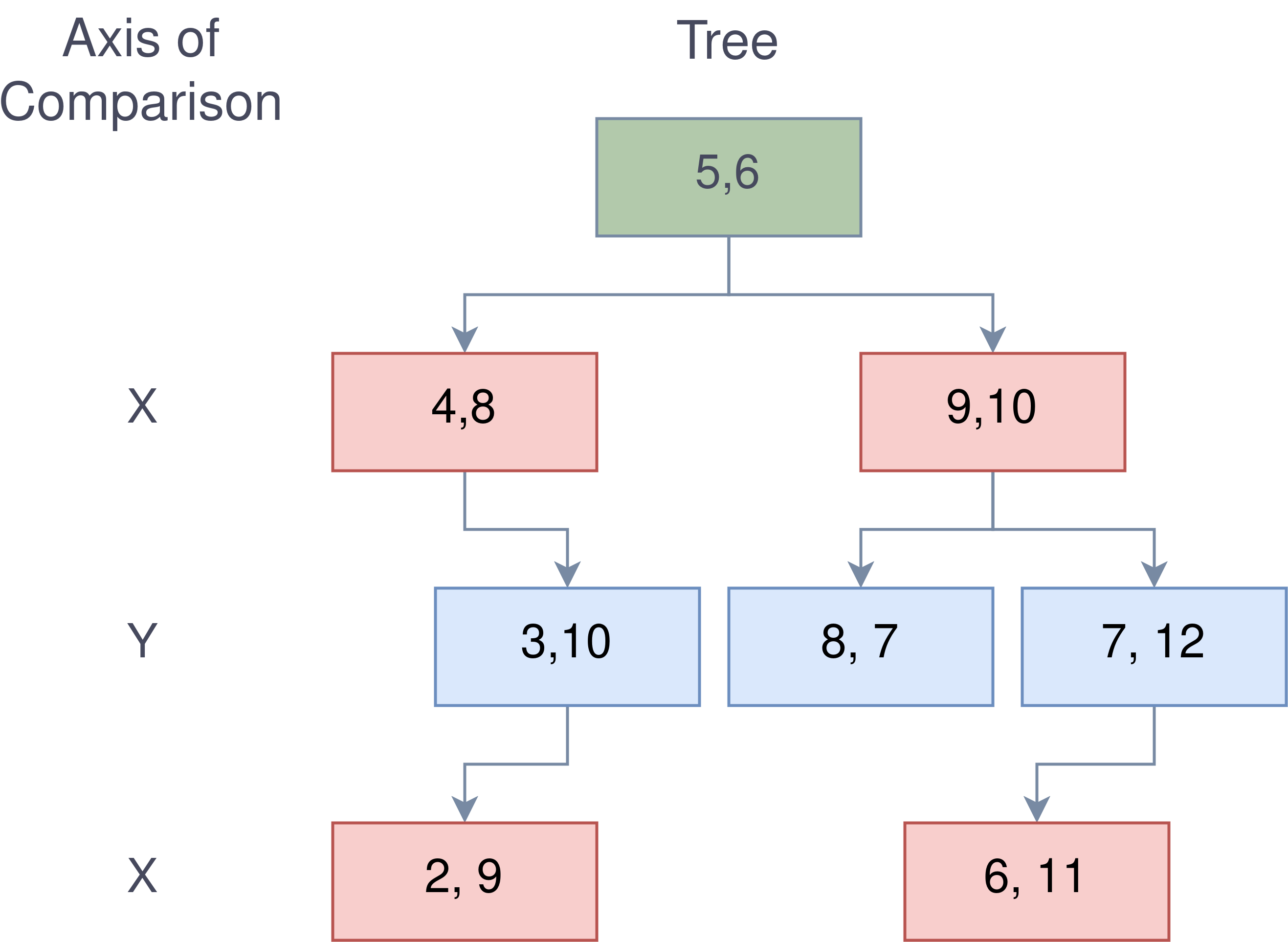}
    \caption{K-d tree partitioning two-dimensional data}
    \label{fig:kd-tree}
\end{figure}

The multidimensional binary search tree, also known as the k-d tree, was proposed as a data structure to store information to be retrieved by associative searches \cite{Bentley1975-dt}.
A k-d tree is a space partitioning data tree where each leaf has a dimensionality of $k$.
Using Fig. \ref{fig:kd-tree} as an example, moving down the tree from the root alternates what feature is being compared.
In the case of Fig. \ref{fig:kd-tree}, the comparison alternates between features $X$ and $Y$ axes. 
Similarly to a binary search tree, values that are less than parent node are stored on the left side of the parent node, and values that are greater than the parent node are stored on the right side.
This allows us have similar runtime performances using k-d trees as the binary search tree of $O(log n)$.

With this efficient data structure in place, CoMTE utilizes our previously trained TCN model and the training data. 
All of the training data is fed to the model to determine how accurate the model is on the training data. 
The predictions from the model are split into two groups: correct and incorrect. 
The incorrectly labeled data points are discarded.
The correctly labeled data points are used to fit a k-d tree per class, in our case two. 
These k-d trees are stored for future counterfactual generation.

\subsubsection{Counterfactual Generation}
The counterfactual generation of CoMTE receives a three inputs: the k-d trees, the data point to be explained and the desired class. 
For a counterfactual query, the specified class would indicate which k-d tree is searched. 
Using their proposed sequential greedy algorithm and a specified number of distractors, a sequential process is done that replaces the feature values of the data point for values of the corresponding feature in the distractor. 
This process is repeated until the data successfully gets labeled as the specified class by our model.
Ates et al. \cite{ates2021counterfactual} also propose a random-restart hill climbing approach for counterfactual generation, but we utilized the greedy approach for our proof of concept. 

%% file: SECTIONS/whatif_method.tex
The last element of our framework involves the interaction between a user and the explainable PdM system. 
We focus on the actual generation of explanations instead of focusing on the interface of the explainable system.
The interface is an equally valuable work that deserves its own attention, but this lies outside of the scope of this work. 

The interaction between user and system is best described through a visual, referencing Figure. \ref{fig:storyboard}.
The system would alert the user of an anomaly.
The user would then be able to query the system for a counterfactual to see what the system would look like if it were healthy. 
This counterfactual could then be used as a way of simulating a what-if analysis as it would be answering the question "what if our system had been performing normally?"
Through this what-if analysis, the user would be able to request an additional counterfactual if that one was unsatisfactory; moreover, the user would be able to request a new counterfactual if the required change would not be possible given the time frame, supplies, or other extraneous factors.

%% file: SECTIONS/results_kfold.tex
\input{SECTIONS/results_kfold_table}
As seen in Table \ref{tab:kfold}, all models perform well, above 98\% accuracy, regardless of the fold and dataset. 
Additionally, the standard deviations of the accuracies are very small, below 0.5. 
This indicates that the model set up that we use would perform well on unseen data.

%% file: SECTIONS/results_kfold_table.tex
\begin{table}[ht]
    \centering
    \begin{tabular}{|c|c c c c c|c|c|c|}
    \hline
        \textbf{Bearing} & & & \textbf{Fold} & & & \textbf{Mean} & \textbf{StD} \\ \cline{2-6}
         & 1& 2& 3& 4& 5& &\\ \hline
         Bearing 1 &98.88 &98.60 &99.02 &99.51 &99.23 &99.04& 0.30\\ \hline
         Bearing 2 &99.70 &99.10 &98.95 &99.55 &99.85 &99.43& 0.34\\ \hline
         Bearing 3 &98.57 &99.40 &99.05 &98.45 &99.40 &98.98& 0.40\\ \hline
    \end{tabular}
    \caption{K-fold anomaly detection accuracy results for the different bearing datasets}
    \label{tab:kfold}
\end{table}

%% file: SECTIONS/results_tcn.tex
\input{SECTIONS/results_tcn_table}
As seen in Table \ref{tab:tcn_performance}, all models performed well in regards to some metric. 
$TCN_1$ performed the best in the most amount of metrics. 
Namely it had the best accuracy on all of the datasets, including being tied with $TCN_2$ and $TCN_3$ on the Bearing3 dataset. 
Additionally, $TCN_1$ had the best precision and f-score on the Bearing1 dataset. 
$TCN_2$ had the best g-score on the Bearing1 dataset.
$TCN_3$ had the best precision, f-score and g-score on the Bearing3 dataset. 
Lastly, $TCN_{123}$ had the best recall on Bearing1 and Bearing3. 

%% file: SECTIONS/results_tcn_table.tex
\begin{table}[ht]
    \centering
\begin{tabular}{|c|c|c|c|c|}
    \hline
                            &                   &\multicolumn{3}{|c|}{\textbf{Test Dataset}} \\ \hline
                           & \textbf{Model}    & \textbf{Bearing1}  & \textbf{Bearing2}  & \textbf{Bearing3}\\ \hline
    \textbf{Accuracy}       & $TCN_1$           & \textbf{0.9977}    & \textbf{0.9991}    & \textbf{0.9574}\\
                            & $TCN_2$           & 0.9818    & 0.9888    & \textbf{0.9574} \\
                            & $TCN_3$           & 0.9685    & 0.9934    & \textbf{0.9574} \\
                            & $TCN_{123}$       & 0.0039    & 0.0       & 0.0426 \\ \hline
    \textbf{Recall}         & $TCN_1$           & 0.8       & ---       & 0.0\\
                            & $TCN_2$           & 0.9714    & ---       & 0.0 \\
                            & $TCN_3$           & 0.9429    & ---       & 0.1333 \\
                            & $TCN_{123}$       & \textbf{1.0}       & ---       & \textbf{1.0} \\ \hline
    \textbf{Precision}      & $TCN_1$           & \textbf{0.6667}    & ---       & 0.0\\
                            & $TCN_2$           & 0.1717    & ---       & 0.0 \\
                            & $TCN_3$           & 0.1044    & ---       & \textbf{0.5} \\
                            & $TCN_{123}$       & 0.0039    & ---       & 0.0426 \\ \hline
    \textbf{F-Score}        & $TCN_1$           & \textbf{0.7272}    & ---       & 0.0\\
                            & $TCN_2$           & 0.2918    & ---       & 0.0 \\
                            & $TCN_3$           & 0.1880    & ---       & \textbf{0.2105} \\
                            & $TCN_{123}$       & 0.0077    & ---       & 0.0817 \\ \hline
    \textbf{G-Score}        & $TCN_1$           & 0.3994    & ---       & 0.0\\
                            & $TCN_2$           & \textbf{0.4769}    & ---       & 0.0 \\
                            & $TCN_3$           & 0.4566    & ---       & \textbf{0.0663} \\
                            & $TCN_{123}$       & 0.0       & ---       & 0.0 \\ \hline
\end{tabular}
    \caption{Performance of different TCN models on the different datasets on a scale of 0-1. Bold text indicates best performance of that metric with that dataset. `---' indicate a non-applicable metric.}
    \label{tab:tcn_performance}
\end{table}

%% file: SECTIONS/discussion.tex
\subsection{K-Fold}
The k-fold experiment was quite successful with all three datasets. 
The accuracies for all of the folds show that the TCN model is good enough for anomaly detection on the PRONOSTIA dataset. 
Additionally, it consistently gets a high accuracy which indicates good generalizability. 
\subsection{TCN for Anomaly Detection}
Firstly, there is a seemingly odd performance with all of the models on the Bearing2 dataset. 
The reason for 0.0 being an overwhelming value through the metrics on this dataset is two-fold.
PRONOSTIA does not have true values for anomaly detection like it does remaining useful life. 
This would create the need to create the labels. 
Leading into the second reason, the three-sigma method utilizing RMS did not label any of the test data in Bearing2 as anomalous. 

Secondly, $TCN_{123}$ showed perfect recall on Bearing1 and Bearing3. 
It would also show perfect recall on Bearing2 if there were any anomalous data in the testing set.
This belief is brought on by it constantly predicting anomaly, which can be seen in the very small accuracy but high recall. 
This model would need to either be trained longer or be more complex to learn what it means to be an ``anomalous bearing" regardless of the operating condition. 

Overall, the models trained on one of the datasets performed well regarding some metrics.
This is even the case with performing well on datasets that were not part of their training data.

\subsection{What-if Analysis}
\begin{figure}
    \includegraphics[width=0.54\textwidth, left]{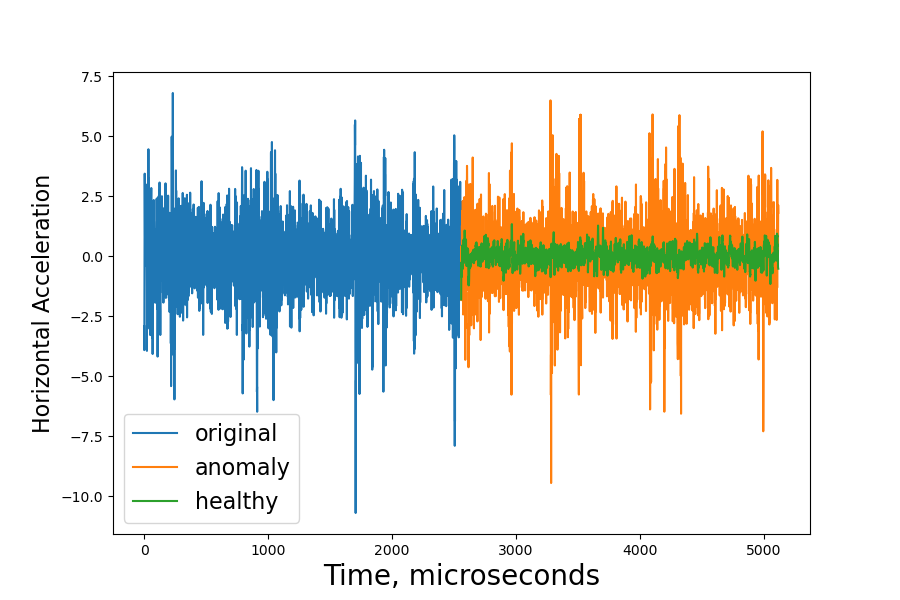}
    \caption{An example of a counterfactual as what-if analysis from Bearing1 test dataset. Blue data represents the time prior to anomaly detection. Orange data represents the actual data where the anomaly is detected. Green data represents the counterfactual that shows how the bearing could be performing if it was performing healthily. 
    }
    \label{fig:whatif_example}
\end{figure}

As for the results of our what-if analysis, we provide an example from the Bearing1 test dataset where the $TCN_1$ model was used as the anomaly detector.
As seen in Fig. \ref{fig:whatif_example}, the original data is shown in blue, the real anomalous data is shown in orange, and the counterfactual is shown in green.
This depicts two separate paths for the life of this bearing, one of anomaly (orange) and one of intervention to perform healthily (green). 
This is helpful as it depicts how the bearing should be acting if the maintainer wanted the bearing to be more associated with healthy data. 
This horizontal acceleration should be slowed to approximately $\pm{2}$. 
Now to answer the question ``what if the bearing had been performing healthily?"
If the bearing had been performing healthily, its acceleration would have been more similar to the counterfactual, e.g. oscillating between $\pm{2}.$
This would also be useful to the maintainer as they would know to put less stress on the bearing in that way specifically which could allow for a longer lifetime of the bearing. 

%% file: SECTIONS/limitation.tex
In terms of limitations and improvements that could be made upon our work, the system is relatively simple to match the dataset. 
There are namely two places that could need alternative methods as the the data scales, the class balancing method and the counterfactual generation approach. 
Both of these approaches may not scale well as the number of features increases. 
For class balancing, methods for synthetic data generation such as Variational Autoencoders and Generative Adversarial Network could be implemented for better performance \cite{iglesias2023data}. 
For counterfactual generation, the k-d tree is very fast when querying it. 
The limitation comes in the creation of the k-d tree. 
The authors propose using k-means clustering to get exemplars for k-d tree creation, but there could still be a desire to keep full dataset to get the best counterfactuals. 
Other approaches could be tested such as the ones presented in \cite{liu2024practical, delaney2021instance,sun2024counterfactual}.
Of course this method could still suffice for larger datasets, and verifying this is just one opportunity for future work.

%% file: SECTIONS/conclusion.tex
Anomaly detection is an important component of predictive maintenance that aims to determine whether or not a component of a cyber-psychical system shows unhealthy behavior. 
If a maintainer of a cyber-physical system, such as a vehicle with a rich array of sensors, was in a maintenance station with a full array of tools, the maintainer would be able to perform any maintenance task.
This is an ideal.
Continuing with the vehicle example, a vehicle in use may be away from many of the tools and parts used for maintenance; moreover, an anomaly would not be so easily fixed. 
What-if analysis allows the maintainer to test different hypotheses in the form of simulations that could point to an elongated lifetime of the system by performing a separate repair, changing their behavior, etc.
Counterfactual explanation generation can provide these simulations by providing realistic examples of healthy data with minimal or specified changes to the system. 
Our experiments showed the Temporal Convolutional Network to be a good classifier for anomaly detection on the PRONOSTIA bearing dataset. 
The Counterfactuals for Multivate Time-series Explanations method showed to be a fast performing method that provided realistic counterfactuals for the anomalous bearing data. 
This work shows to be a useful proof-of-concept that can serve as the basis for more complex environments where what-if analysis will show to be more useful in the field of predictive maintenance. 